\documentclass[runningheads]{llncs}
\usepackage[T1]{fontenc}
\usepackage{graphicx}

\usepackage{multirow}
\usepackage[dvipsnames]{xcolor}
\usepackage{wrapfig}
\usepackage{numprint}
\usepackage{url}

\begin{document}
\npdecimalsign{.}
\nprounddigits{3}

\title{M3D-NCA: Robust 3D Segmentation with Built-in Quality Control}
%
%
\author{John Kalkhof\orcidID{0000-0001-7316-1903}\thanks{Corresponding author: john.kalkhof@gris.tu-darmstadt.de, +49 6151 155-681} \and 
Anirban Mukhopadhyay\orcidID{0000-0003-0669-4018}} 

%
\authorrunning{J. Kalkhof et al.}
%
\institute{Darmstadt University of Technology, Karolinenplatz 5, 64289 Darmstadt, Germany}
\maketitle              
\begin{abstract}

Medical image segmentation relies heavily on large-scale deep learning models, such as UNet-based architectures. However, the real-world utility of such models is limited by their high computational requirements, which makes them impractical for resource-constrained environments such as primary care facilities and conflict zones. Furthermore, shifts in the imaging domain can render these models ineffective and even compromise patient safety if such errors go undetected. To address these challenges, we propose M3D-NCA, a novel methodology that leverages Neural Cellular Automata (NCA) segmentation for 3D medical images using \textit{n-level} patchification. Moreover, we exploit the variance in M3D-NCA to develop a novel quality metric which can automatically detect errors in the segmentation process of NCAs. M3D-NCA outperforms the two magnitudes larger UNet models in hippocampus and prostate segmentation by 2\% Dice and can be run on a Raspberry Pi 4 Model B (2GB RAM). This highlights the potential of M3D-NCA as an effective and efficient alternative for medical image segmentation in resource-constrained environments.

\end{abstract}
\section{Introduction}
Medical image segmentation is ruled by large machine learning models which require substantial infrastructure to be executed. These are variations of UNet-style \cite{UNet} architectures that win numerous grand challenges \cite{nnUnet}. This emerging trend raises concerns, as the utilization of such models is limited to scenarios with abundant resources, posing barriers to adoption in resource-limited settings. For example, conflict zones \cite{jaff2019improving}, low-income countries \cite{frija2021improve}, and primary care facilities in rural areas \cite{resourceConstrained_overview} often lack the necessary infrastructure to support the deployment of these models, impeding access to critical medical services. Even when the infrastructure is in place, shifts in domains can cause the performance of deployed models to deteriorate, posing a risk to patient treatment decisions. To address this risk, automated quality control is essential \cite{gonzalez2022distance}, but it can be difficult and computationally expensive.

Neural Cellular Automata (NCA) \cite{gilpin2019cellular} diverges strongly from most deep learning architectures. Inspired by cell communication, NCAs are one-cell models that communicate only with their direct neighbours. By iterating over each cell of an image, these relatively simple models, with often sizes of less than \emph{13k parameters}, can reach complex global targets. By contrast, UNet-style models quickly reach \emph{30m parameters} \cite{kalkhof2023med}, limiting their area of application. Though several minimal UNet-style architectures with backbones such as EfficientNet \cite{efficientnet}, MobileNetV2 \cite{mobilenetv2}, ResNet18 \cite{resnet} or VGG11 \cite{vgg11} exist, their performance is generally restricted by their limited size and still require several million parameters.

\begin{figure}[t]
  \centering
  \includegraphics[width=.97\linewidth]{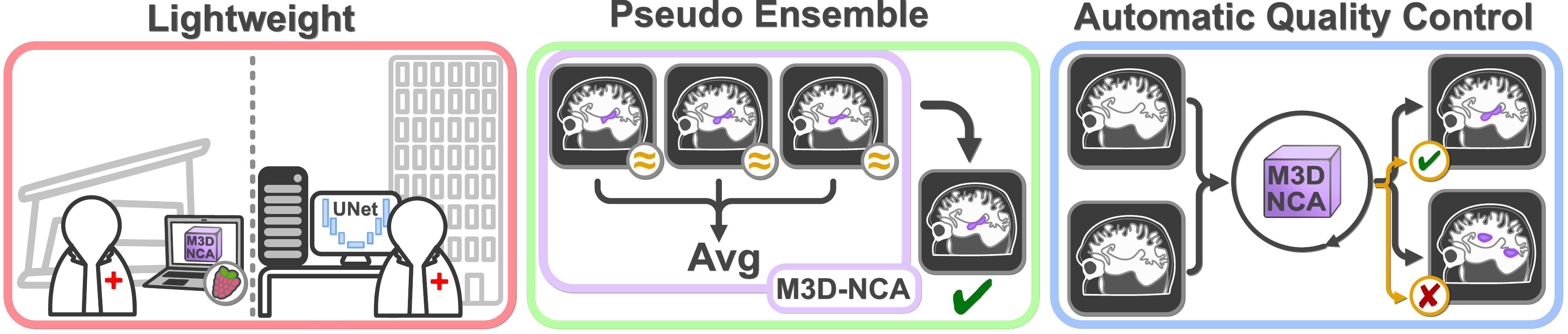}
  \caption{M3D-NCA is lightweight, with a parameter count of less than 13k and can be run on a Raspberry Pi 4 Model B (2GB RAM). The stochasticity enables a pseudo-ensemble effect that improves prediction performance. This variance also allows the calculation of a score that indicates the quality of the predictions.}
  \label{fig:IntroductionGraphic}
\end{figure}

With Med-NCA, Kalkhof et al. \cite{kalkhof2023med} have shown that by iterating over two scales of the same image, high-resolution 2D medical image segmentation using NCAs is possible while reaching similar performance to UNet-style architectures. While this is a step in the right direction, the limitation to two-dimensional data and the fixed number of downscaling layers make this method inapplicable for many medical imaging scenarios and ultimately restricts its potential.

Naively adapting Med-NCA for three-dimensional inputs exponentially increases VRAM usage and \emph{convergence becomes unstable}. We address these challenges with M3D-NCA, which takes NCA medical image segmentation to the third dimension and is illustrated in Fig. \ref{fig:IntroductionGraphic}. Our \textbf{n-level architecture} addresses VRAM limitations by training on patches that are precisely adaptable to the dataset requirements. Due to the one-cell architecture of NCAs the \textbf{inference can be performed on the full-frame image}. Our \emph{batch duplication scheme} stabilizes the loss across segmentation levels, enabling segmentation of high-resolution 3D volumes with NCAs. In addition, we propose a \emph{pseudo-ensemble} technique that exploits the stochasticity of NCAs to generate multiple valid segmentations masks that, when averaged, improve performance by 0.5-1.3\%. Moreover, by calculating the variance of these segmentations we obtain a quality assessment of the derived segmentation mask. Our \emph{NCA quality metric (NQM)} detects between 50\% (prostate) and 94.6\% (hippocampus) of failure cases. M3D-NCA is \emph{lightweigth} enough to be run on a Raspberry Pi 4 Model B (2GB RAM).

We compare our proposed M3D-NCA against the UNet \cite{UNet}, minimal variations of UNet, Seg-NCA \cite{sandler2020image} and Med-NCA \cite{kalkhof2023med} on the medical segmentation decathlon \cite{medicalsegmentationdecathlon} datasets for hippocampus and prostate. M3D-NCA consistently outperforms minimal UNet-style and other NCA architectures by at least 2.2\% and 1.1\% on the hippocampus and prostate, respectively, while being at least \emph{two magnitudes smaller} than UNet-style models. However, the performance is still lower than the nnUNet by 0.6\% and 6.3\% Dice, the state-of-the-art auto ML pipeline for many medical image segmentation tasks. This could be due to the additional pre-and post-processing steps of the pipeline, as well as the extensive augmentation operations.

We make our complete framework available under \url{github.com/MECLabTUDA/M3D-NCA}, including the trained M3D-NCA models for both anatomies as they are only \emph{56KB} in size.
 
\begin{figure}[t]
  \centering
  \includegraphics[width=.85\linewidth]{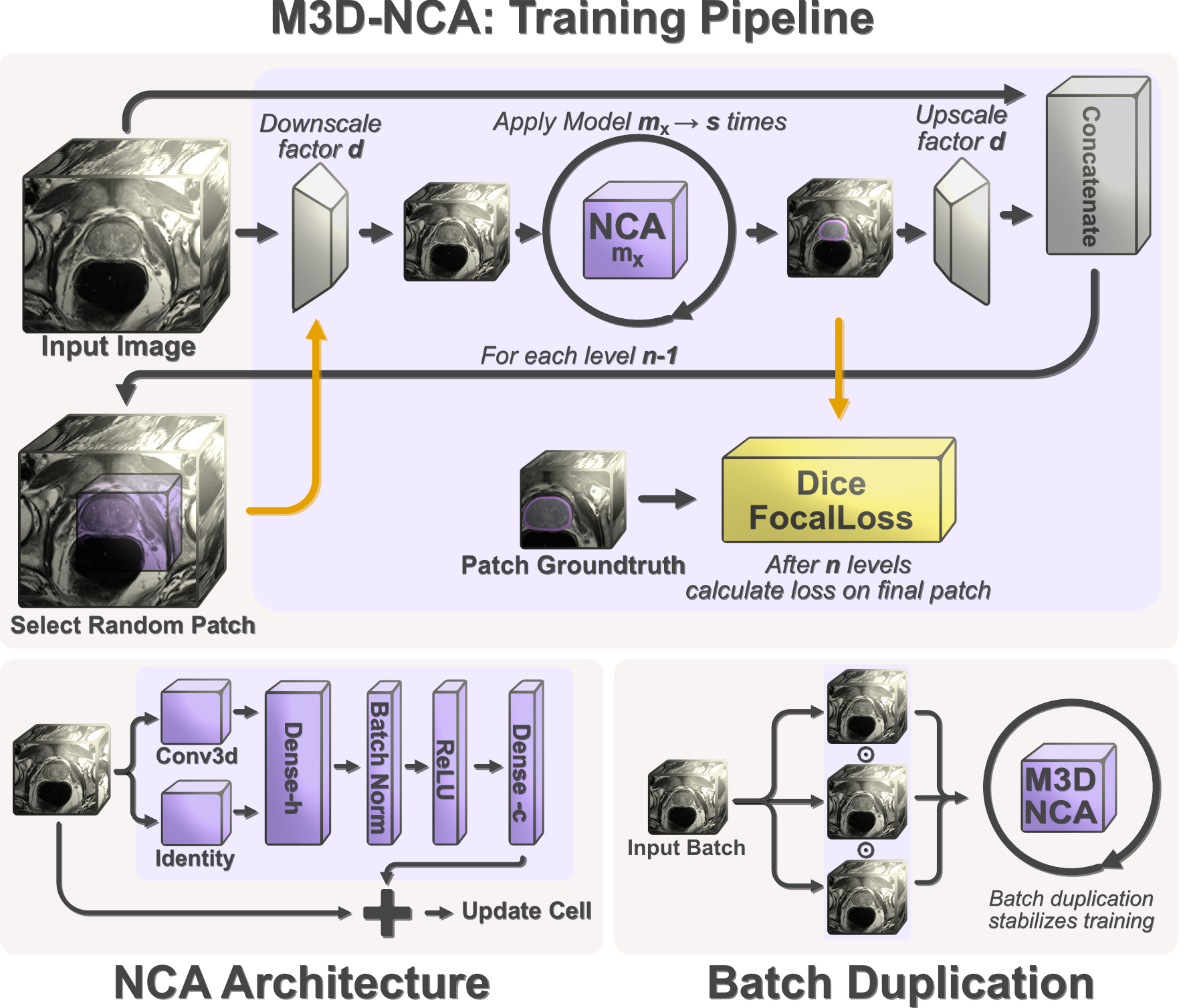}
  \caption{The n-level M3D-NCA architecture uses \emph{patchification} and \emph{batch duplication} during training.}
  \label{fig:Med3NCA}
\end{figure}

\section{Methodology}

Cellular Automata (CA) are sets of typically hand-designed rules that are iteratively applied to each cell of a grid. They have been actively researched for decades, \emph{Game Of Life} \cite{gardner1970fantastic} being the most prominent example of them. Recently, this idea has been adapted by Gilpin et al. \cite{gilpin2019cellular} to use neural networks as a representation of the update rule. These \textbf{Neural Cellular Automata} (NCA) are minimal and interact only locally (illustration of a 2D example can be found in the supplementary). Recent research has demonstrated the applicability of NCAs to many different domains, including image generation tasks \cite{growingNCA}, self-classification \cite{randazzo2020self}, and even 2D medical image segmentation \cite{kalkhof2023med}.

NCA segmentation in medical images faces the problem of high VRAM consumption during training. Our proposed M3D-NCA described in Sec. \ref{sec:Med3NCA} solves this problem by performing segmentation on different scales of the image and using patches during training. In Sec. \ref{sec:inherentQuality} we introduce a score that indicates segmentation quality by utilizing the variance of NCAs during inference.

\subsection{M3D-NCA Training Pipeline}
\label{sec:Med3NCA}
Our core design principle for M3D-NCA is to minimize the VRAM requirements. Images larger than $100 \times 100$, can quickly exceed 40 GB of VRAM, using a naive implementation of NCA, especially for three-dimensional configurations.

The training of M3D-NCA operates on different scales of the input image where the same model architecture $m$ is applied, as illustrated in Fig. \ref{fig:Med3NCA}. The input image is first downscaled by the factor $d$ multiplied by the number of layers $n$. If we consider a setup with an input size of $320 \times 320 \times 24$, a downscale factor of $d=2$, and $n=3$, the image is downscaled to $40 \times 40 \times 3$. As $d$ and $n$ exponentially decrease the image size, big images become manageable. On this smallest scale, our first NCA model $m_1$, which is constructed from our core architecture (Section \ref{sec:core}), is iterated over for $s$ steps, initializing the segmentation on the smallest scale. The output of this model gets upscaled by factor $d$ and appended with the according higher resolution image patch. Then, a random patch is selected of size $40 \times 40 \times 3$, which the next model $m_2$ iterates over another $s$ times. We repeat this patchification step $n-1$ times until we reach the level with the highest resolution. We then perform the dice focal loss over the last remaining patch and the according ground truth patch. Changing the downscaling factor $d$ and the number of layers $n$ allows us to precisely control the VRAM required for training.

\label{sec:MirroredInputs}
\textbf{Batch Duplication:} Training NCA models is inherently more unstable than classical machine learning models like the UNet, due to two main factors. First, stochastic cell activation can result in significant jumps in the loss trajectory, especially in the beginning of the training. Second, patchification in M3D-NCA can cause serious fluctuations in the loss function, especially with three or more layers, thus it may never converge properly. 

The solution to this problem is to duplicate the batch input, meaning that the same input images are multiple times in each batch. While this limits the number of images per stack, it greatly improves convergence stability.

\begin{figure}[htbp]
  \centering
  \includegraphics[width=.85\linewidth]{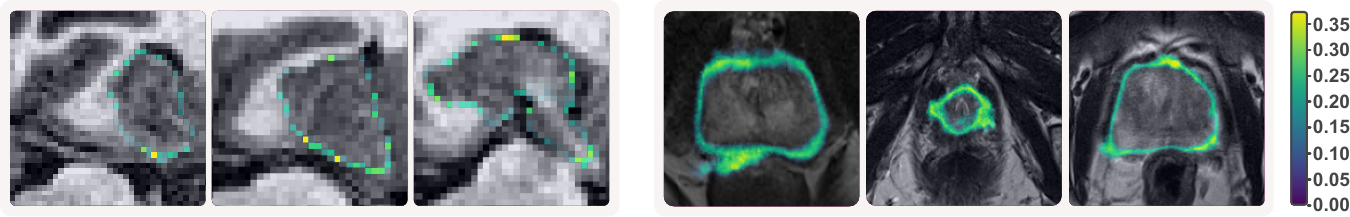}
  \caption{Variance over 10 predictions on different samples of the hippocampus (left) and prostate dataset (right).}
  \label{fig:pseudo_ensemble}
\end{figure}

\textbf{Pseudo Ensemble:} The stochasticity of NCAs, caused by the random activation of cells gives them an inherent way of predicting multiple valid segmentation masks. We utilize this property by executing the trained model 10 times on the same data sample and then averaging over the outputs. We visualize the variance between several predictions in Fig. \ref{fig:pseudo_ensemble}.

Once the model is trained, inference can be performed directly on the full-scale image. This is possible due to the one-cell architecture of NCAs, which allows them to be replicated across any image size, even after training.

\subsection{M3D-NCA Core Architecture}
\label{sec:core}

The core architecture of M3D-NCA is optimized for simplicity. First, a convolution with a kernel size $k$ is performed, which is appended with the identity of the current cell state of depth $c$ resulting in state vector $v$ of length $2*c$. $v$ thus contains information about the surrounding cells and the knowledge stored in the cell. $v$ is then passed into a dense layer of size $h$, followed by a 3D BatchNorm layer and a ReLU. In the last step, another Dense layer is applied, which has the output size $c$, resulting in the output being of the same size as the input. Now the cell update can be performed, which adds the model's output to the previous state. Performing a full execution of the model requires it to be applied $s$ times. In the standard configuration, the core NCA sets the hyperparameters to $k=7$ for the first layer, and $k=3$ for all the following ones. $c=16$ and $h=64$ results in a model size of $12480$ parameters. The bigger k in the first level allows the model to detect low-frequency features, and $c$ and $h$ are chosen to limit VRAM requirements. The steps $s$ are determined per level by $s=max(width,height,depth)/((k-1)/2)$, allowing the model to communicate once across the whole image.

\subsection{Inherent Quality Control}
\label{sec:inherentQuality}
The variance observed in the derived segmentation masks serves as a quantifiable indicator of the predicted segmentation. We expect that a higher variance value indicates data that is further away from our training domain and consequently may lead to poorer segmentation accuracy. Nevertheless, relying solely on this number is problematic, as the score obtained is affected by the size of the segmentation mask. To address this issue, we normalize the metric by dividing the sum of the standard deviation by the number of segmentation pixels.

The \emph{NCA quality metric (NQM)} where $v$ is an image volume and $v_i$ are $N=10$ different predictions of M3D-NCA for $v$ is defined as follows:

\begin{equation}
NQM = \frac{\sum_{s \in SD}(s)}{\sum_{m \in \mu}(m)}, \;\;\; SD = \sqrt{\frac{\sum_{i = 1}^{N}(v_i - \mu)^2}{N}}, \;\;\; \mu = \frac{\sum_{i = 1}^{N}v_i}{N}
\end{equation} 

We calculate the relation between Dice and NQM by running a linear regression on the training dataset, which has been enriched with spike artifacts to extend the variance range. Using the regression, we derive the detection threshold for a given Dice value (e.g., $Dice > 0.8$). In clinical practice, this value would be based on the task and utility.

\begin{figure}[t]
  \centering
  \includegraphics[width=.85\linewidth]{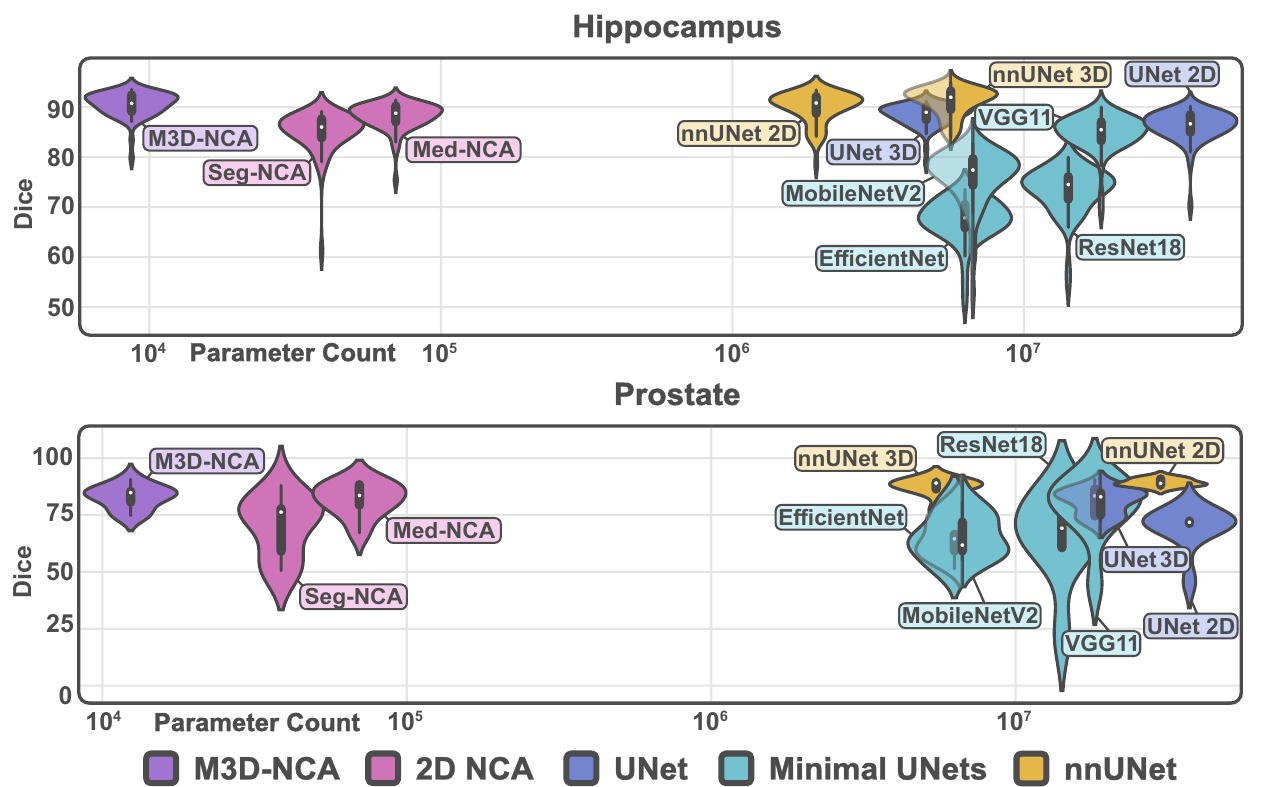}
  \caption{Comparison of the Dice segmentation performance versus the number of parameters of NCA architectures, minimal UNets and the nnUNet (check supplementary for detailed numbers).}
  \label{fig:ModelPerformance}
\end{figure}

\section{Experimental Results}
The evaluation of the proposed M3D-NCA and baselines is performed on hippocampus (198 patients, $\sim 35 \times 50 \times 35$) and prostate (32 patients, $\sim 320 \times 320 \times 20$) datasets from the medical segmentation decathlon (\url{medicaldecathlon.com}) \cite{medicalsegmentationdecathlon,simpson2019large}. All experiments use the same 70\% training, and 30\% test split and are trained on an \textit{Nvidia RTX 3090Ti} and an \textit{Intel Core i7-12700}. We use the standard configuration of the \textit{UNet} \cite{fernando_perez_garcia_2019_3522306}, \textit{Segmentation Models Pytorch} \cite{segmentation_models_pytorch} and \textit{nnUNet} \cite{isensee2021nnu} packages for the implementation in PyTorch \cite{paszke2019pytorch}.

\begin{figure}[htbp]
  \centering
  \includegraphics[width=.85\linewidth]{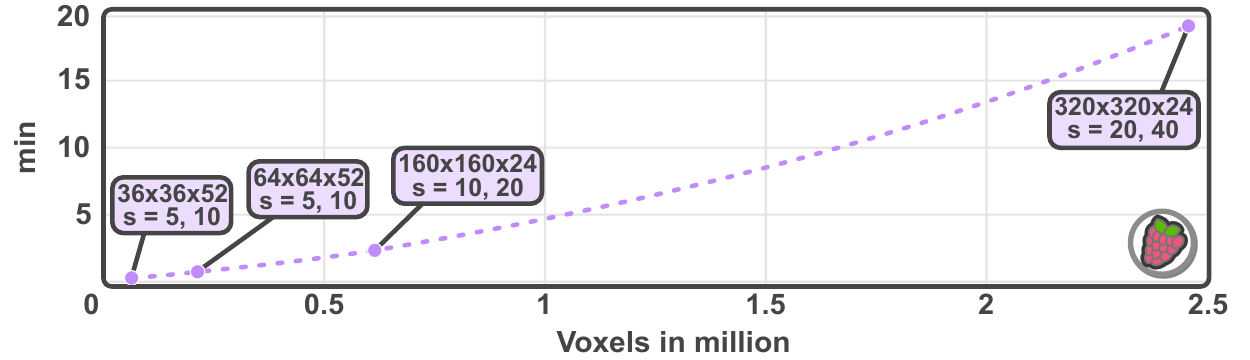}
  \caption{Example inference times of a 2-level M3D-NCA architecture across different image scales on a \emph{Raspberry Pi 4 Model B (2GB RAM)}, where $s$ defines the number of steps in each layer.}
  \label{fig:InferenceTime}
\end{figure}
 
 \subsection{Comparison and Ablation}
Our results in Fig. \ref{fig:ModelPerformance} show that despite their compactness, M3D-NCA performs comparably to much larger UNet models. UNet-style models instead tend to underperform when parameter constraints are imposed. While an advanced training strategy, such as the auto ML pipeline nnUNet, can alleviate this problem, it involves millions of parameters and requires a minimum of 4GB of VRAM \cite{nnUnet}.

In contrast, our proposed M3D-NCA uses \textbf{two orders of magnitude fewer parameters}, reaching 90.5\% and 82.9\% Dice for hippocampus and prostate respectively. M3D-NCA outperforms all basic UNet-style models, falling short of the nnUNet by only 0.6\% for hippocampus and 6.3\% for prostate segmentation. Utilizing the 3D patient data enables M3D-NCA to outperform the 2D segmentation model Med-NCA in both cases by 2.4\% and 1.1\% Dice. The Seg-NCA \cite{sandler2020image} is due to its one-level architecture limited to small input images of the size 64x64, which for prostate results in a performance difference of 12.8\% to our proposed M3D-NCA and 5.4\% for hippocampus. We execute M3D-NCA on a Raspberry Pi 4 Model B (2GB RAM) to demonstrate its suitability on resource-constrained systems, as shown in Fig. \ref{fig:InferenceTime}. Although our complete setup can be run on the Raspberry Pi 4, considerably larger images that exceed the device's 2GB memory limit require further optimizations within the inference process. By asynchronously updating patches of the full image with an overlapping margin of $(k-1)/2$ we can circumvent this limitation while ensuring identical inference.

\begin{table}
\centering
\begin{tabular}{|c|c|c|c|c|c|} 
\hline
\multirow{2}{*}{\textbf{Lay.}} & \multirow{2}{*}{\textbf{Scale F.}} & \multirow{2}{*}{ \textbf{\# Param.} $\downarrow$ } & \textbf{Standard} & \textbf{ w/o Batch Dup.} & \textbf{ w/o Pseudo E.}  \\ 
&&& \textbf{Dice} $\uparrow$ & \textbf{Dice} $\uparrow$ & \textbf{Dice} $\uparrow$  \\ 
\hline
2 & 4 & 12480 & $\mathbf{\numprint{0.8291528158717685}\pm\numprint{0.0507199596148773}}$ & $\mathbf{\numprint{0.8110266725222269}\pm\numprint{0.04477488931720998}}$ & $\mathbf{\numprint{0.8236347647}\pm\numprint{0.05115609057}}$\\ 
3 & 2 & 16192 & $\numprint{0.8021034465895759}\pm\numprint{0.03826854169814999}$ & $\numprint{0.7227533525890775}\pm\numprint{0.1030374913244215}$ & $\numprint{0.7893621233}\pm\numprint{0.0407351808}$\\
4 & 2 & \textbf{8880} & $\numprint{0.7466674115922716}\pm\numprint{0.11238729530939928}$ & $\numprint{0.7039483984311422}\pm\numprint{0.21098039177553193}$ & $\numprint{0.7337668962}\pm\numprint{0.1165446213}$\\
\hline
\end{tabular}
\newline
\caption{Ablation results of M3D-NCA on the prostate dataset.}
\label{Tab:QuantHP}
\end{table}

The ablation study of M3D-NCA in Tab. \ref{Tab:QuantHP} shows the importance of batch duplication during training, especially for larger numbers of layers. Without batch duplication, performance drops by 1.8-7.9\% Dice. Increasing the number of layers reduces VRAM requirements for larger datasets, but comes with a trade-off where each additional layer reduces segmentation performance by 2.7-5.5\% (with the 4-layer setup, a kernel size of 5 is used on the first level, otherwise the downscaled image would be too small). The pseudo-ensemble setup improves the performance of our models by 0.5-1.3\% and makes the results more stable. 

\begin{figure}[htbp]
  \centering
  \includegraphics[width=.85\linewidth]{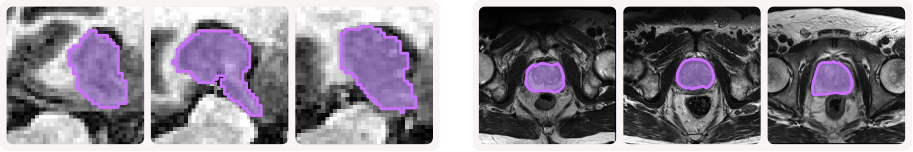}
  \caption{Qualitative Results of M3D-NCA on hippocampus (left) and prostate (right).}
  \label{fig:qualitative}
\end{figure}

The qualitative evaluation of M3D-NCA, illustrated in Fig. \ref{fig:qualitative}, shows that M3D-NCA produces accurate segmentations characterized by well-defined boundaries with no gaps or random pixels within the segmentation volume.

\subsection{Automatic Quality Control}
\begin{figure}[htbp]
  \centering
  \includegraphics[width=.85\linewidth]{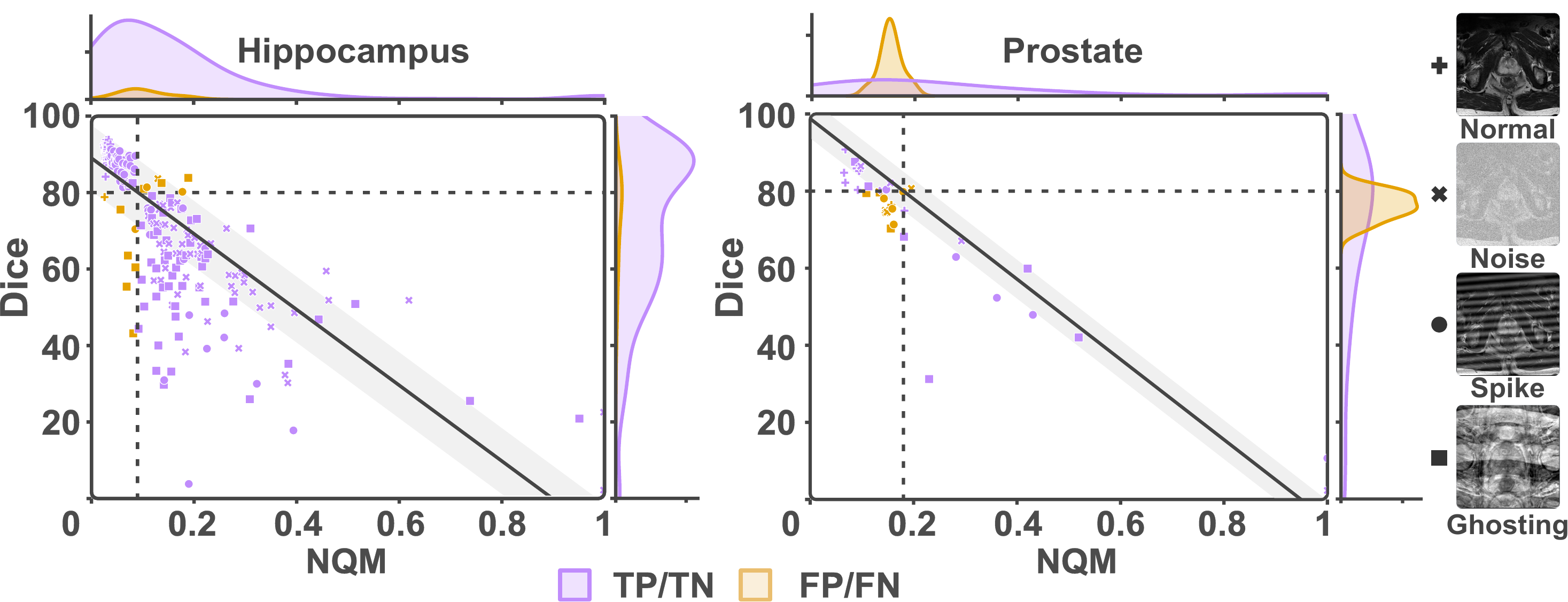}
  \caption{The variance of NCAs during inference encapsulated in the NQM score indicates the quality of segmentation masks. In this example, the calculated threshold should detect predictions worse than 80\% Dice. The distribution of FP/FN cases shows that most fall close to the threshold.}
  \label{fig:CertaintyPrediction}
\end{figure}

To evaluate how well M3D-NCA identifies failure cases through the NQM metric, we degrade the test data with artifacts using the \textit{TorchIO} package \cite{perez-garcia_torchio_2021}. More precisely, we use noise ($std=0.5$), spike ($intensity=5$) and ghosting ($num\_ghosts=6$ and $intensity=2.5$) artifacts to force the model to collapse (prediction / metric pairs can be found in the supplementary). We effectively identify 94.6\% and 50\% of failure cases (below 80\% Dice) for hippocampus and prostate segmentation, respectively, as shown in Fig. \ref{fig:CertaintyPrediction}. Although not all failure cases are identified for prostate, most false positives fall close to the threshold. Furthermore, the false negative rates of 4.6\% (hippocampus) and 8.3\% (prostate), highlights its value in identifying particularly poor segmentations.

\section{Conclusion}
We introduce M3D-NCA, a Neural Cellular Automata-based training pipeline for achieving high-quality 3D segmentation. Due to the small model size with under 13k parameters, M3D-NCA can be run on a Raspberry Pi 4 Model B (2GB RAM). M3D-NCA solves the VRAM requirements for 3D inputs and the training instability issues that come along. In addition, we propose an NCA quality metric (NQM) that leverages the stochasticity of M3D-NCA to detect 50-94.6\% of failure cases without additional overhead. Despite its small size, M3D-NCA outperforms UNet-style models and the 2D Med-NCA by 2\% Dice on both datasets. This highlights the potential of M3D-NCAs for utilization in primary care facilities and conflict zones as a viable lightweight alternative.

\bibliographystyle{splncs04}
\bibliography{refs}

\end{document}


\npdecimalsign{.}
\nprounddigits{3}

\begin{figure}[!htbp]
  \centering
  \includegraphics[width=.5\linewidth]{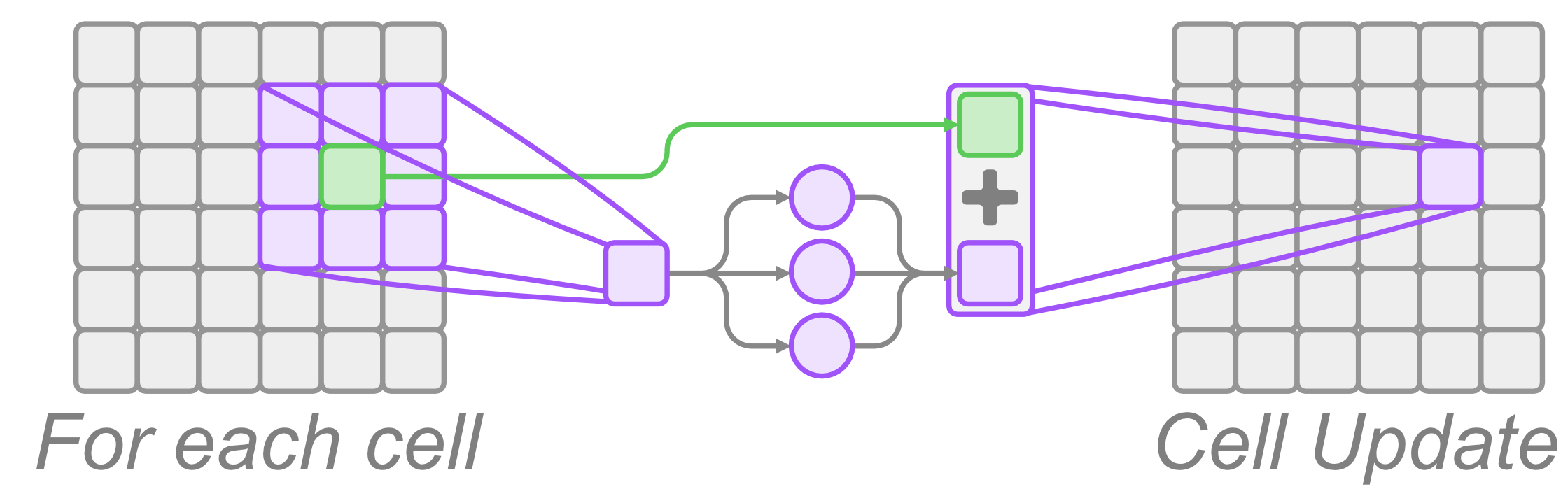}
  \caption{Simplified example of the update of a single cell in a 2D grid using NCAs.}
  \label{fig:simplNCA}
\end{figure}

\begin{table}[!htbp]
\centering
\begin{tabular}{|c|c|c|c|c|}
\hline
\multirow{2}{*}{\textbf{Model}} & \multicolumn{2}{c|}{\textbf{Hippocampus}} & \multicolumn{2}{c|}{\textbf{Prostate}} \\
 & \textbf{Dice} $\uparrow$ & \textbf{\# Parameters} $\downarrow$ & \textbf{Dice} $\uparrow$ & \textbf{\# Parameters} $\downarrow$ \\
\hline
nnUNet 2D & $\numprint{0.8993015525252978}\pm\numprint{0.030995057965490264}$ & 1928032 & $\mathbf{\numprint{0.8919956473122801}\pm\numprint{0.01633238697485009}}$ & 29966112 \\
nnUNet 3D & $\mathbf{\numprint{0.9109540658429306}\pm\numprint{0.02779603167969324}}$ & 5602720 & $\numprint{0.8723066758714417}\pm\numprint{0.03922772838392659}$ & 5471648 \\
\hline
MobileNetV2 & $\numprint{0.759842961627844}\pm\numprint{0.05551850082047271}$ & 6628369 & $\numprint{0.6598262389500936}\pm\numprint{0.08683071013822083}$ & 6628369 \\
EfficientNet & $\numprint{0.6758332404039674}\pm\numprint{0.04093023995919301}$ & 6250893 & $\numprint{0.6448938912815518}\pm\numprint{0.09308752357877809}$ & 6250893\\
ResNet18 & $\numprint{0.733060663029299}\pm\numprint{0.046226724441606616}$ & 14321937 & $\numprint{0.6628889044125875}\pm\numprint{0.17784321638298703}$ & 14321937\\
VGG11 & $\numprint{0.8487855226306592}\pm\numprint{0.036007825930041745}$ & 18252881 & $\numprint{0.7816272113058302}\pm\numprint{0.1307224044184063}$ & 18252881\\
\hline
UNet 3D & $\numprint{0.8834458023814832}\pm\numprint{0.022140445258470212}$ & 4584769 & $\numprint{0.8077938093079461}\pm\numprint{0.052470459465187656}$ & 19071297 \\ 
UNet 2D & $\numprint{0.8623192471972967}\pm\numprint{0.030953540741727}$ & 36950273 & $\numprint{0.6940711935361227}\pm\numprint{0.085685931421014}$ & 36950273 \\
Seg-NCA & $\numprint{0.8507985745446157}\pm\numprint{0.04455198220948615}$ & 39472 & $\numprint{0.7024800247616239}\pm\numprint{0.1259253604424353}$ & 39472 \\ 
Med-NCA & $\numprint{0.8811472866494777}\pm\numprint{0.02700984083121783}$ & 70016 & $\numprint{0.8184222380320231}\pm\numprint{0.06924227997981384}$ & 70016 \\
M3D-NCA & $\mathbf{\numprint{0.9053496273897462}\pm\numprint{0.0244843732685336}}$ & \textbf{8768} & $\mathbf{\numprint{0.8291528158717685}\pm\numprint{0.0507199596148773}}$ & \textbf{12480} \\
\hline
\end{tabular}
\newline
\caption{Comparison of our proposed M3D-NCA with state-of-the-art NCA and UNet segmentation models.}
\label{Tab:QuantHP}
\end{table}

\begin{figure}[!htbp]
  \centering
  \includegraphics[width=.91\linewidth]{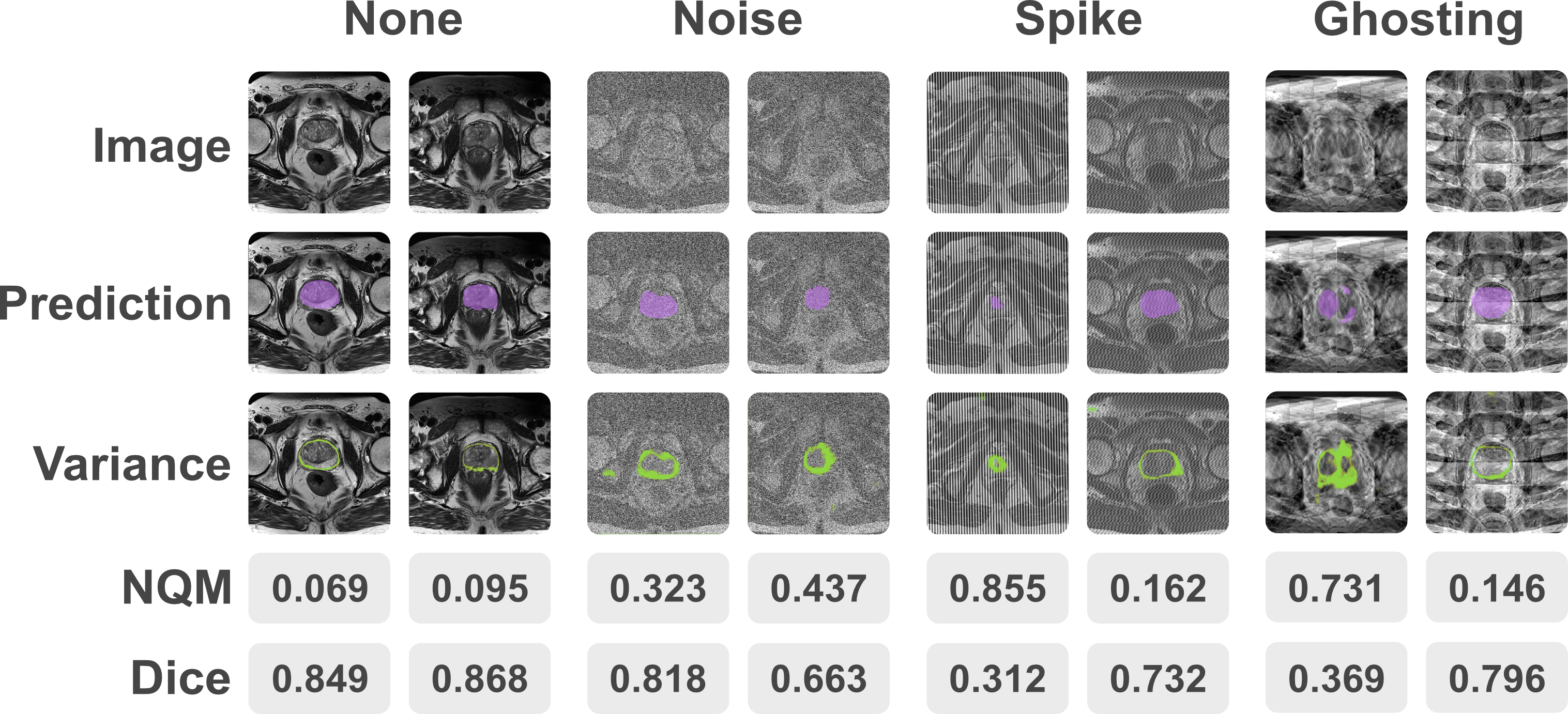}
  \caption{Comparison between the prediction of M3D-NCA, the variance between predictions, the calculated NQM score and the Dice on the prostate dataset.}
  \label{fig:CompNQMDice}
\end{figure}